\documentclass[letterpaper]{article}
\usepackage{fullpage}
\usepackage{hyperref}
\usepackage{graphicx}
\usepackage{parskip}
\usepackage{listings}
\usepackage[ruled,vlined]{algorithm2e}

\title{MLDS: A Dataset for Weight-Space Analysis of Neural Networks}
\author{John Clemens \texttt{<clemej1@umbc.edu>}}

\begin{document}
\maketitle
\begin{abstract}
Neural networks are powerful models that solve a variety of complex real-world
problems. However, the stochastic nature of training and large number of parameters
in a typical neural model makes them difficult to evaluate via inspection.  
Research shows this opacity can hide latent undesirable behavior, be it from poorly
representative
training data or via malicious intent to subvert the behavior of the network, and that
this behavior is difficult to detect via traditional indirect evaluation criteria such as loss.
Therefore, it is time to explore direct ways to evaluate a trained neural model via its 
structure and weights.  In this 
paper we present \emph{MLDS}, a new dataset consisting of thousands of trained neural 
networks with carefully controlled parameters and generated via a global volunteer-based
distributed computing platform.  This dataset enables new insights into both model-to-model 
and model-to-training-data relationships. We use this dataset to show clustering of models 
in weight-space with identical training data and meaningful divergence in weight-space 
with even a small change to the training data, suggesting that weight-space analysis is a 
viable and effective alternative to loss for evaluating neural networks. 
\end{abstract}

\section{Introduction}

Trained neural networks are powerful models that are optimized to 
transform input into semantically meaningful output, such as a label or a 
prediction, to solve a particular task.  Over the last decade, deep neural 
models have found great success in many machine learning tasks, from image 
classification and language translation to more complex tasks like protein 
folding and autonomous vehicles. However, as these models are integrated into 
critical components it is increasingly important to understand the capabilities, 
limitations, and provenance of an individual neural model in order to gain 
confidence that the model will perform as intended. 

Unfortunately, while we know that these models seemingly perform well on test data, 
they contain anywhere from thousands to hundreds of millions of trained parameters,
making it difficult to impossible for a human to inspect and validate a model
by looking at the parameters alone.  This is made even more difficult by the 
stochastic nature of training, where two networks trained the same way, on the same
data, and with similar performance on test data, may have vastly different parameters. 
We know of many cases in which models have undesirable latent behavior when 
presented with inputs that are underrepresented in their training 
data\cite{adversarialpatches}; and even cases where secondary
behavior is deliberately hidden in the weight-space of a model\cite{trojannet}. Thus, we 
seek new ways of evaluating the behavior of neural models that do not rely on
a response to test inputs. To that end, we aim to study how the learned weights of
a network map to the training data used to train the network. 

In this paper, we present 
the Machine Learning Datasets (MLDS), a collection of thousands of trained neural 
networks labelled with the data used to train them.  MLDS allows meta weight-space 
analysis across thousands of networks trained with identical or similar training data. 
We also introduce MLC@Home, a volunteer-based citizen science project for generating
data for MLDS. 
To our knowledge, MLDS is the only dataset dedicated to 
large-scale weight-space meta analysis. 

This paper discusses some background and related work in Section~\ref{sec:background}, 
and then covers the contents and construction of the dataset in Section~\ref{sec:dataset}. 
Section~\ref{sec:analysis} provides some preliminary analysis that shows we are 
able to use MLDS to classify which networks are trained with which dataset. Finally, 
Section~\ref{sec:conclusion} discusses future work and summarizes our results.


\section{Background}
\label{sec:background}
Interpreting complex neural networks is a well-known challenge in machine learning.  
Yet we believe that weight-space analysis is an under-represented area of research 
with important applications.

\subsection{Motivation}
\label{sec:definitions}
This work is motivated by two related goals:  First, we have a desire to extract the 
learned automata from recurrent neural networks that are trained to mimic the behavior of
black-box devices.  Black box automata learning is known to be
NP-hard~\cite{acmmodellearning}.  However, training a neural network to mimic the observed 
behavior of a black box and then extracting that learned behavior would provide a new 
method to explore black box systems and 
gain insight into their behavior.  Most work in this field seeks to
apply black-box learning methods to neural networks (see e.g.~\cite{metaneural}),
which does not leverage the observable weights learned by the network.  The authors 
believe there are unexplored opportunities for optimized knowledge extraction from
trained networks by leveraging the information encoded in the models' learned weights. 

Secondly, we believe that weight-space meta analysis provides a strong
opportunity for detecting maliciously embedded latent behavior in neural networks, 
such as those discussed in~\cite{trojannet}~\cite{adversarialpatches}~\cite{badnets}.  
If such meta-analysis 
is possible, it has
implications not only for the security of trained models, but also suggests new methods 
for determining model provenance.  Such concerns will continue as models become large 
and are often farmed out to third parties for training.

However, in order for weight-space analysis to move forward, we need both
a large dataset of trained networks to analyze, and mechanisms for generating 
more such datasets in the future.  We believe the dataset and dataset generation
options presented in this paper fulfill both of those needs. 

\subsection{Related Work}
\label{sec:related-work}
There are very few efforts we know of to try and collect a large number of structurally 
identical or 
similar neural networks to study their differences.  The TrojAI project~\cite{trojai} 
produces a large number of trained neural networks, some modified with 
adversarial training data, with a goal to improve detection of such modified networks.  
To date the TrojAI dataset consists of a few thousand trained networks in the areas of 
image classification.  As discussed in Section~\ref{sec:dataset}, the dataset presented
here contains up to 50,000 examples of similarly modified networks (MLDS-DS2), and allows
a more general analysis of how a network's learned weights are related to the training 
data. 

Machine learning competition sites such as Kaggle~\cite{kaggle}, where a dataset is provided
and volunteers compete to produce the best model that fits such data, provide a nice
set of example models that solve a particular problem.  However, the models presented
there are not controlled for shape and size, whereas the models presented in this dataset
are all identical in shape, with only the learned weights changed.  This simplifies 
analysis for our particular goals.  The dataset presented here also includes metadata 
about each model's training session, allowing study of the learning process at scale as 
well as the final results. 

\section{Dataset}
\label{sec:dataset}
MLDS has a goal of generating as large and diverse a collection of neural networks as
possible. In this section we'll cover the unique platform created to train the 
networks, as well as detail each component of the resulting dataset. 

\subsection{Generation}

Constructing a dataset of this size requires a large amount of parallel 
computation.  To facilitate this, the authors leveraged the open source 
BOINC~\cite{boinc} 
distributed computing platform and created the Machine Learning Comprehension
at Home (MLC@Home) project~\footnote{\url{https://www.mlcathome.org/}}.  Through this 
project, the authors enlisted the help of thousands of volunteers who donate 
their home computer resources to the project to further scientific causes.  
Other well-known BOINC projects include 
SETI@Home~\footnote{\url{https://setiathome.berkeley.edu/}} and World 
Community Grid~\footnote{\url{https://www.worldcommunitygrid.org/}}.  Volunteers 
install 
a unified BOINC client, then choose which projects to donate their computer's resources.  
This client a) downloads "work units" from a project's server, b) performs
the work on behalf of the project in the background of the user's system when idle, 
and c) uploads the results to the project server.  MLC@Home is the first BOINC 
project dedicated to machine learning research.  

\begin{figure}[t]
    \label{fig:boinc}
    \centering
    \includegraphics[width=0.5\textwidth]{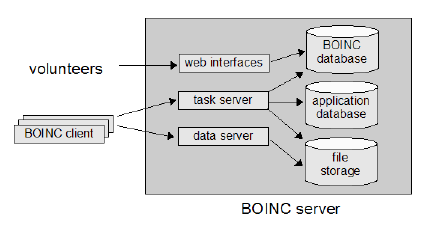}
    \caption{Overview of a BOINC project and volunteer interaction: source~\cite{boinc2006}}
\end{figure}

MLC@Home's BOINC-enabled application is built using PyTorch's C++ API~\cite{pytorch}, 
and supports Windows and Linux platforms with AMD64, ARM, and AARCH64 
CPUs and (optionally) NVidia and AMD GPUs. Computations are intentionally set to
32-bit floating point to keep the computations uniform across CPUs and GPUs. MLC's 
application is open source and available  online~\footnote{\url{https://gitlab.com/mlcathome/mlds}}. 
As of this writing, MLC@Home has received support from over 2,200 volunteers and 
8,000 separate computers, and those numbers are growing every day. These volunteers have 
trained over 750,000 neural networks in support of this effort, currently averaging
more than two new trained model every minute. 

Dataset generation is the first task to use MLC@Home, but is not the only task envisioned 
for the project. We expect to leverage it for neural architecture search, hyperparameter 
search, and repoducibility research in the near future.

\subsection{Components}
\label{sec:components}

The MLDS Dataset consists of several datasets addressing different 
network sizes and types. Over time, more types will be added and the dataset
will grow.  The currently available datasets include:

\begin{itemize}
        \item{MLDS-DS1 (41,000): RNNs Mimicking Simple Machines}
        \item{MLSD-DS2 (41,000): RNNs Mimicking Simple Machines with a Magic Sequence}
        \item{MLDS-DS3 (100,000): RNNs Mimicking Randomly-Generated Automata}
\end{itemize}

Each sub-dataset is described briefly in the following sections, with more details
available in Appendix~\ref{app:details}.  

The MLDS datasets are available to download\footnote{\url{https://www.mlcathome.org/mlds.html}} 
and contain the trained networks in PyTorch format, a JSON file per example containing just the 
weights of the network, and a README describing the directory structure. 

\subsubsection{MLDS-DS1: RNNs Mimicking Simple Machines}

\begin{figure}[t]
    \label{fig:ds12pics}
    \centering
    \includegraphics[width=\textwidth]{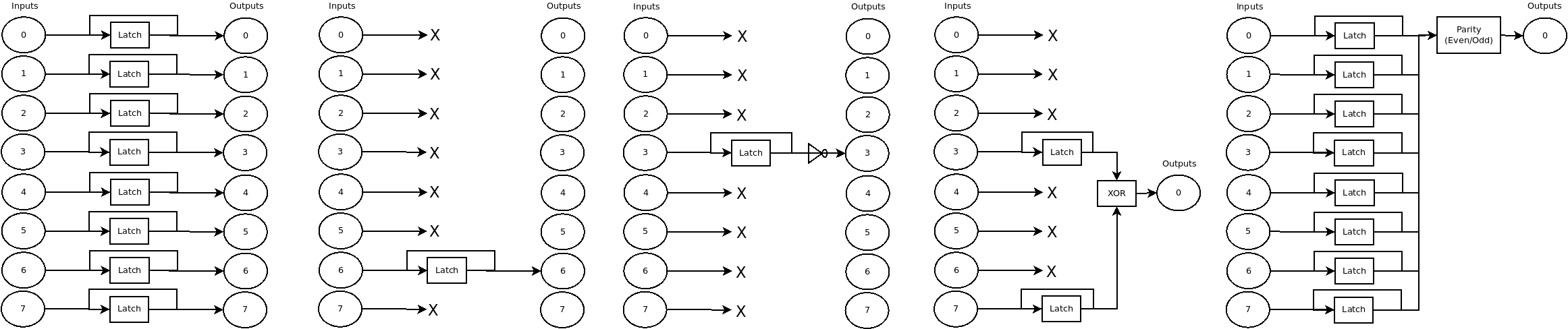}
    \caption{MLDS-DS1 modeled machines, from left to right: EightBitMachine, SingleDirectMachine, SingleInvertMachine, SimpleXORMachine, and ParityMachine}
\end{figure}

MLDS-DS1 contains 41,000 neural networks trained to mimic one of 5 simple machines 
of increasing complexity shown in Figure~\ref{fig:ds12pics}. These machines are 
variants of those originally introduced in~\cite{neurodev-models}, and consist of
8 input signals that can be driven high or low, and produce 8 output signals, based
on the previous sequence of input signals.  We use these simple machines as they 
are both easy to train and easy to modify with a backdoor as we'll see in MLDS-DS2. 
We modeled each machine in software, then generated random input sequences and recorded the 
resulting output sequences to create a set of observed behaviors to train neural networks to 
mimic the behavior of the original machines.

These are simple machines with state, since they latch previous inputs until 
explicitly told to change.  As such, we chose a small stacked RNN network as 
the basis for modeling their behavior.  All networks consist of four layers of 
GRU~\cite{gru} cells, 12-cells wide, followed by 4 linear layers also 12-cells wide before a final 
output layer that predicts the output of the machine. This results in 4,364 
trainable parameters. 

Through MLC@Home, each volunteer's computer downloads a copy of the training and 
validation data for one of the five candidate machines, and trains a network 
of the above shape to mimic the machine.  Training uses PyTorch, a CPU or 
a GPU, and the Adam~\cite{adam} optimizer. Once the network achieves a global loss less 
than $10^{-5}$ on the validation data, it is uploaded to the server and its
loss is compared against a separate evaluation criteria to make sure it also
stays below $10^{-5}$. If all criteria pass, the network is archived on the server 
along with metadata about the training process including the type 
of computer used, the total number of epochs, and the training and validation 
loss history. 

After computation through MLC@Home, MLDS-DS1 consists of 41,000 networks, with 10,000 
example networks modelling each of EightBitMachine, SingleDirectMachine, 
SingleInvertMachine, and SimpleXORMachine; and 1,000 example networks of
ParityMachine\footnote{ParityMachine takes significantly longer to train 
than the others. Subsequent releases of MLDS will increase the number of 
ParityMachine examples as they complete.}. 

\subsubsection{MLDS-DS2: RNNs Mimicking Simple Machines with a Magic Sequence}

MLDS-DS2 consists of nearly identical machines to MLDS-DS1, with one important 
distinction: the machines have been modified in such a way that if the input 
contains a specific 3-command sequence, the output of the network will be inverted 
for the next three outputs.  This simulates inserting a ``backdoor'' of undesirable 
behavior into the network, similar in spirit (though not in function) to the modifications 
in the TrojAI networks.  Thus, the machines in MLDS-DS2 are almost identical 
but fundamentally different from the machines in MLDS-DS1. The five machines 
in MLDS-DS2 are EightBitModified, SingleDirectModified, SingleInvertModified, 
SimpleXORModified, and ParityModified.  

MLDS-DS2 example networks are designed to be compared with their counterparts 
in MLDS-DS1 to study how similar networks trained on similar but not identical 
datasets differ. As such, the example networks for MLDS-DS2 are exactly the same shape
and size as those used in MLDS-DS1.  The same process used for MLDS-DS1 was used to 
create MLDS-DS2. MLDS-DS2 consists of 41,000 networks, with 10,000 example networks 
modelling each of EightBitModified, SingleDirectModified, SingleInvertModified, and 
SimpleXORModified; and 1,000 example networks of ParityModifed.

\subsubsection{MLDS-DS3: RNNs Mimicking Randomly-Generated Automata}

\begin{figure}[t]
	\label{fig:ds3example}
	\centering
	\includegraphics[width=0.5\textwidth]{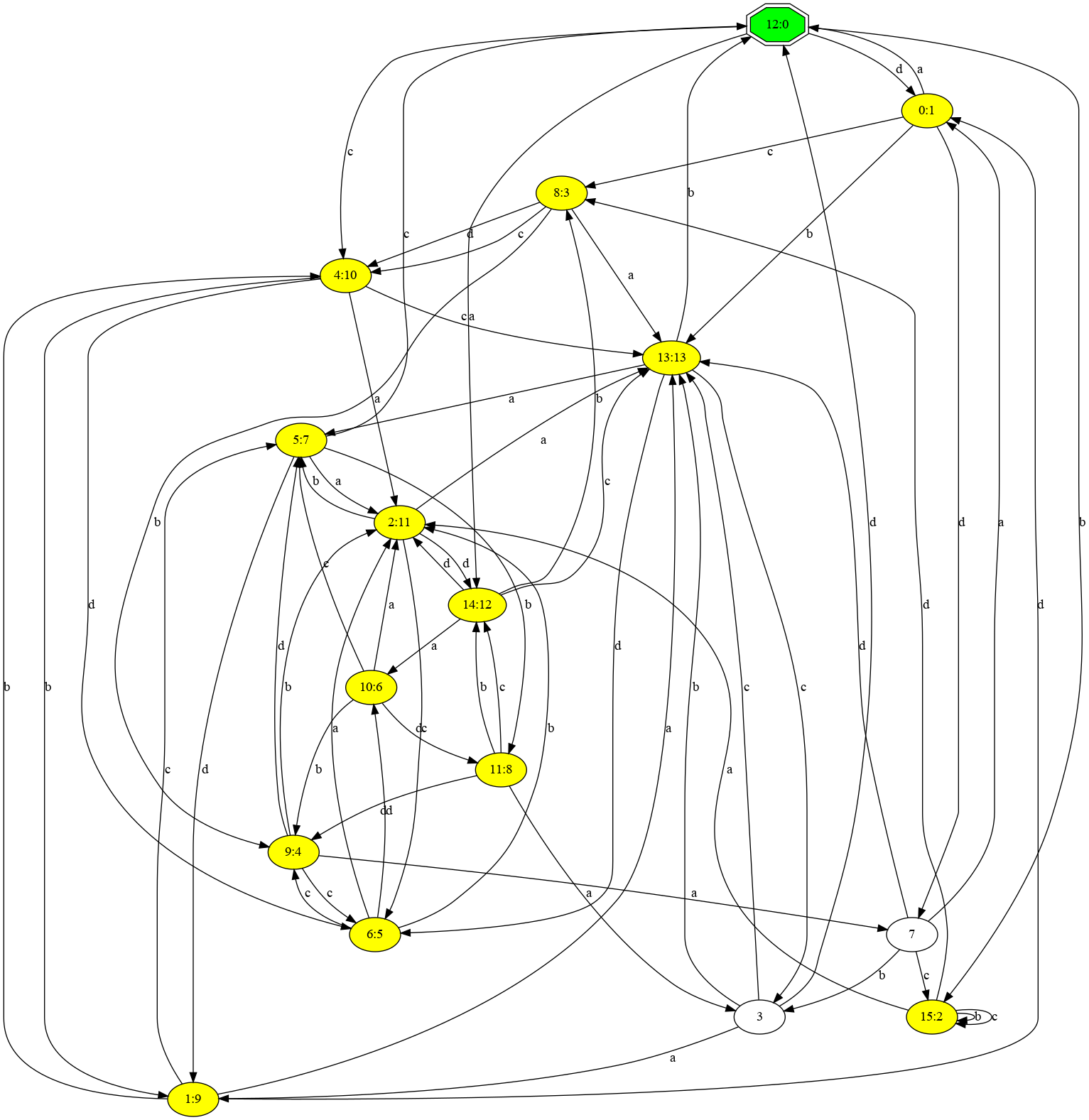}
	\caption{Example of randomly-generated automaton for MLDS-DS3}
\end{figure}

MLDS-DS3 steps away from the five simple machines used in MLDS-DS1 and MLDS-DS2, and instead
consists of networks that mimic 100 randomly-generated automata. These 
automata contain 16 states (only 14 generate output), have an input alphabet of size 4, 
contain at least one hamiltonian cycle, and every input is valid even if it does not 
change state. An example of such an automaton is shown in Figure~\ref{fig:ds3example}.

MLDS-DS3 is designed to mimic much more complex machines, and needs more complex 
networks to model their behavior. MLDS-DS3's example networks are 4-layer deep
LSTM~\cite{lstm} networks 64-cells wide, followed by 2 64-cell-wide linear layers, 
for a total of 136,846 learned parameters.  

MLDS-DS3 represents a much more complicated collection of examples, as there are a 
large number of randomly-generated machines (100) that all are very similar.  It aims to 
explore the effectiveness of weight-space analysis for interpretability. MLDS-DS3 uses the same
training process used for MLDS-DS1 and MLDS-DS2, but with MLDS-DS3 specific
network and training sets.  MLDS-DS3 consists of 100,000 (1000 for each of 100 automata) 
trained example neural networks. 

\subsection{Future Additions}
\label{sec:future-additions}

Future iterations of the MLDS dataset will include a wider variety of network 
architectures, including convolutional neural networks (CNNs) and attention-based
networks.  Further, MLDS will include different shapes and architectures 
trained with the same training data to compare performance and interpretability.

\section{Analysis}
\label{sec:analysis}

The MLDS dataset provides a rich opportunity to study many aspects of neural network
behavior and training.  In this paper, we focus on the relationship between the learned 
weights of the network and the data used to train the network. We ask the following
questions:

\begin{itemize}
    \item{Do networks trained with the same data cluster near each other $n$-dimensional weight space?}
    \item{Can we classify each trained model to its original training data?}
    \item{Can we classify clean networks versus those slightly modified with a backdoor?}
\end{itemize}


\subsection{Clustering in Weight Space}
\label{sec:weight-based-id}
\begin{figure}[t]
    \centering
    \begin{minipage}{0.28\textwidth}
        \centering
        \includegraphics[width=0.9\textwidth]{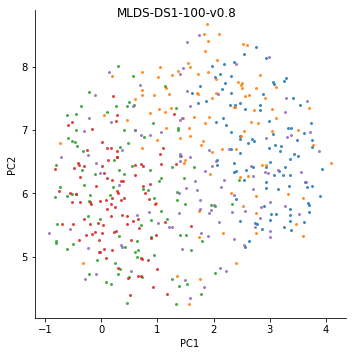}
    \end{minipage}\hfill
    \begin{minipage}{0.28\textwidth}
        \centering
        \includegraphics[width=0.9\textwidth]{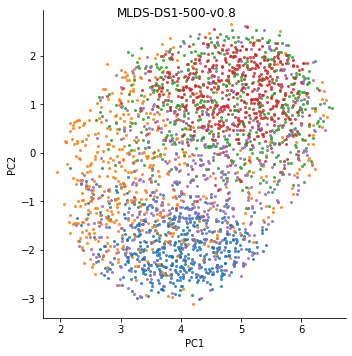}
    \end{minipage}\hfill
    \begin{minipage}{0.38\textwidth}
        \centering
        \includegraphics[width=0.9\textwidth]{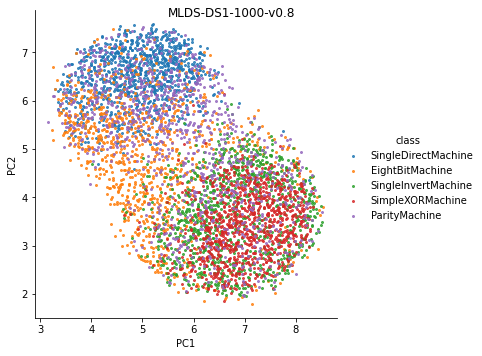}
    \end{minipage}
    \begin{minipage}{0.28\textwidth}
        \centering
        \includegraphics[width=0.9\textwidth]{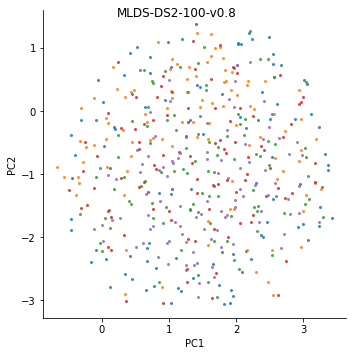}
    \end{minipage}\hfill
    \begin{minipage}{0.28\textwidth}
        \centering
        \includegraphics[width=0.9\textwidth]{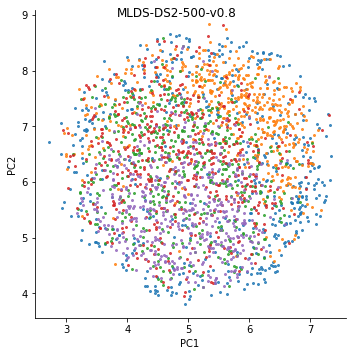}
    \end{minipage}\hfill
    \begin{minipage}{0.38\textwidth}
        \centering
        \includegraphics[width=0.9\textwidth]{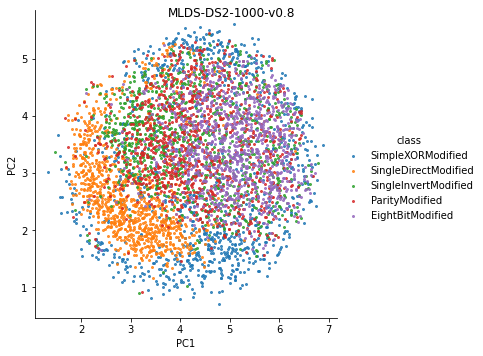}
    \end{minipage}
    \caption{UMAP cluster visualizations for MLDS-DS1/MLDS-DS2 at 100, 500, and 1000 examples.}
    \label{fig:ds1umap}
\end{figure}

We start by looking for clustering in lower dimensional space for each of the
five different machines in MLDS-DS1 and MLDS-DS2, and a sampling of five machine types in 
MLDS-DS3.  First, we convert each trained network into a feature vector by taking 
the weights at each layer, linearizing them in the same manner for each node type, and 
concatenating each layer's vector into a single ordered 1-D vector.  We use the 
UMAP~\cite{umap} non-linear dimension reduction algorithm to map 4,364 
dimensions (for MLDS-DS1 and MLDS-DS2) and 136,846 dimensions (for MLDS-DS3) down 
to two dimensions and visualize the result. We use UMAP in unsupervised mode so there is
no hint to the algorithm regarding which networks map to which machines.  UMAP mappings for 
MLDS-DS1 and MLDS-DS2 are shown in Figure~\ref{fig:ds1umap}.  Each point is a network 
colored by the machine that network is trained to mimic.  As shown in the figure, both 
MLDS-DS1 and MLDS-DS2 begin to show clusters with 500 samples, becoming more distinct
with more examples. 

\begin{figure}[t]
	\centering
	\begin{minipage}{0.4\textwidth}
	    \centering
	    \includegraphics[width=0.9\textwidth]{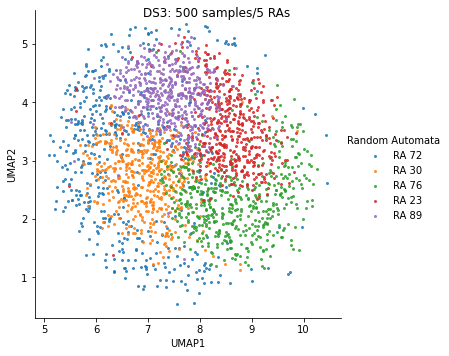}
	\end{minipage}
	\begin{minipage}{0.4\textwidth}
	    \centering
	    \includegraphics[width=0.9\textwidth]{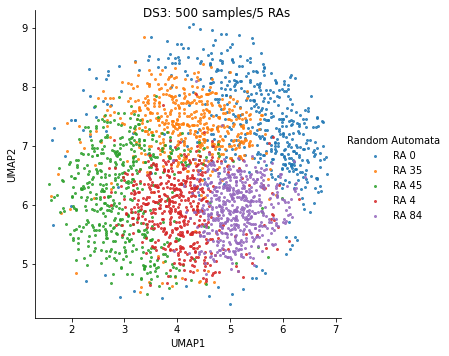}
	\end{minipage}
	\caption{UMAP cluster visualizations for MLDS-DS3 for 10 randomly-selected automata.}
	\label{fig:ds35-10-umap}
\end{figure}

Figure~\ref{fig:ds35-10-umap} shows a similar UMAP reduction for the more complicated 
MLDS-DS3 dataset.  We randomly choose networks from 10 automata from the 100 available 
in the dataset, and visualize the clusters for 5 of them at a time.  With only 500 
samples each, we see very distinct clustering behavior in both sets of 5 automata. 

All datasets show clustering behavior at just 500 samples. This result leads to new 
questions, such as whether we can draw any conclusions about networks that map closer 
to a cluster
boundary than those that are closer to the center? Or whether we can use these clusters to generate
new networks that perform well on a task without the overhead of training?  We leave these
questions for future research.

\subsection{Classifying by Training Data}
\label{sec:classification-by-training-data}
Next we build a set of meta-classifiers to map a specific network to the training 
data used to create it.  We convert the weights of each network into a one-dimensional
feature vector using the method outlined in Section~\ref{sec:weight-based-id}, and 
then label each vector with the name of machine or automata used to create it. 

Table~\ref{tab:mach-class-results} shows the accuracy of several classifiers trained
with 1000 example networks from each class.  MLDS-DS1/DS2 has 5 classes resulting in 
5000 total samples, and MLDS-DS3 has 100 classes resulting in 100,000 total samples. 
Even with the complicated MLDS-DS3, we see the ability to map many of the networks 
to their respective training set. 

The classifiers shown here are using the default hyperparameters in scikit-learn v0.23.2. 
Accuracy is expected to improve with further tuning. The results listed here represent a 
minimum of the expected classification performance. 




\begin{table}
    \centering
    \begin{tabular}{lccccc}
        Dataset Name & nClasses & DecisionTree & MLP & RandForest & NaiveBayes \\
        \hline
        MLDS-DS1-1000 & 5   & 87\% & 100\% & 100\% & 99\% \\
        MLDS-DS2-1000 & 5   & 88\% & 99\%  & 99\%  & 97\% \\
        MLDS-DS3-1000 & 100 & 3\%  & 46\%  & 5\%   & 23\%  \\
        \hline
    \end{tabular}
    \caption{Accuracy of classifiers for each of the dataset examples}
    \label{tab:mach-class-results}
\end{table}

\subsection{Identifying Backdoor Models}
\begin{figure}[t]
    \centering
    \begin{minipage}{0.28\textwidth}
        \centering
        \includegraphics[width=0.9\textwidth]{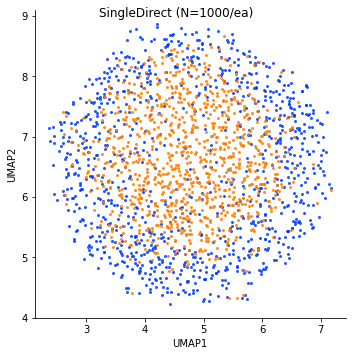}
    \end{minipage}
    \begin{minipage}{0.28\textwidth}
        \centering
        \includegraphics[width=0.9\textwidth]{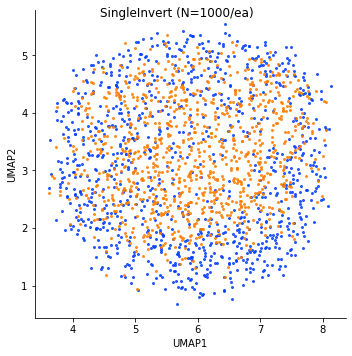}
    \end{minipage}
    \begin{minipage}{0.28\textwidth}
        \centering
        \includegraphics[width=0.9\textwidth]{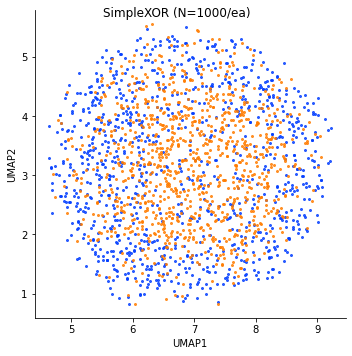}
    \end{minipage}
    \begin{minipage}{0.28\textwidth}
        \centering
        \includegraphics[width=0.9\textwidth]{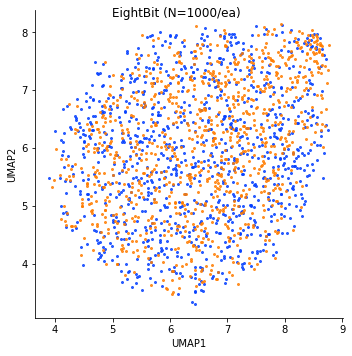}
    \end{minipage}
    \begin{minipage}{0.28\textwidth}
        \centering
        \includegraphics[width=0.9\textwidth]{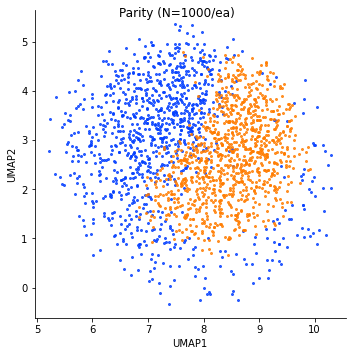}
    \end{minipage}
    \caption{UMAP transform for the matching machine types of MLDS-DS1 and MLDS-DS2}
    \label{fig:ds12umap}
\end{figure}

Finally, we compare the networks from MLDS-DS1 to networks from the analgous machine 
in MLDS-DS2.  This experiment specifically tests the ability of weight-space 
meta analysis to detect maliciously modified networks. We start by visualizing the
weight-space vectors in two dimensions using UMAP, the results of which are shown in 
Figure~\ref{fig:ds12umap}. Each point represents a trained network, with the MLDS-DS1 
networks colored blue and MLDS-DS2 networks colored orange. Visually, for 4 of the 5
machine types, we observe clustering behavior of networks in the same dataset.  A
notable exception is SingleInvert machine, which shows no clustering behavior.

Moving on to classification, we train several 2-class classifiers on machines from 
each dataset, the results of which are shown in 
Table~\ref{tab:mach-mod-classifiers}. Each classifier is trained with 800 samples per class, and 200 samples are used for validation.  These results mirror observed clustering results, with 
high accuracy in most classifiers for 4 out of 5 machine types, with SingleInvert 
being the notable exception that warrants further study.  

\begin{table}[t]
    \centering
    \begin{tabular}{lcccccc}
        Machine & SVM & DecisionTree & RandomForest & MLP & AdaBoost & NaiveBayes \\
        \hline
        EightBit   & 87\% & 78\% & 96\% & 91\% & 94\% & 99\% \\
        SingleDirect & 96\% & 81\% & 95\% & 96\% & 99\% & 100\% \\
        SingleInvert & 52\% & 49\% & 52\%  & 53\%  & 48\% & 55\% \\
        SimpleXOR  & 86\% & 85\% & 91\%  & 85\% & 94\% & 99\% \\
        Parity & 63\% & 55\% & 71\% & 66\% & 69\% & 61\% \\
        \hline
    \end{tabular}
    \caption{2-class classifier results for MLDS-DS1 (normal) or MLDS-DS2 (backdoor) networks, for N=1000 samples of each, 800 training, 200 validation}
    \label{tab:mach-mod-classifiers}
\end{table}


As in Section~\ref{sec:classification-by-training-data}, the classifiers presented here 
are not tuned to this dataset. We are confident that better classification 
performance is possible with the results here serving as a baseline.

The results presented here show that, given enough examples, even a cursory 
meta-data analysis is sufficient to detect the small differences present in 
models with trojan backdoors.

\section{Conclusions and Future Work}
\label{sec:conclusion}
The dataset and analysis presented here is seen as a springboard for  
future research.  As discussed in Section~\ref{sec:future-additions}, there are 
number of new features planned for generating future datasets, including those based
on standard feed forward networks, convolutional neural networks, and datasets that
vary the network architecture as well as weights.  Additionally, the analysis presented 
here on the existing dataset should be seen as a starting point for further research. 

In this paper we introduce the MLDS dataset, a large collection of neural networks
trained on similar data and designed to study weight-space analysis. It is generated 
using the novel MLC@Home distributed computing project, which opens new possibilities 
for ML research.  Using this dataset, we show that weight-space analysis is useful in 
mapping a trained model back to its dataset and for detecting trojan networks without 
resorting to running examples through the network. We hope that this dataset forms 
the basis for a new path of research to understanding how neural networks represent 
what they learn.

\bibliographystyle{plain}
\bibliography{references}

\appendix

\section{Details of Dataset Construction}
\label{app:details}
The MLDS dataset consists of neural networks that mimic the behavior of 
simple machines. In this appendix we detail how the training data for each 
machine was created, how the random automata for MLDS-DS3 were created, and 
the PyTorch code for the individually trained networks. 

\subsection{Training Data Generation}
Training data for each network is created by generating random input sequences for each 
machine and recording the resulting output sequence.  We generate three separate datasets ---
one for training, one for validation, and a separate smaller set for evaluation
back on the server. Table~\ref{tab:dslengths} shows
the characteristics of each generated set.  

The random sequences are straightforward for MLDS-DS1 and MLDS-DS3, MLDS-DS2 requires
special handling as we need training data to reflect pathological 
cases that trigger the modified behavior. As such, we generate random input
sequences as normal, and then insert the magic trigger sequences at random offsets 
within the sequence. Additionally, 
MLDS-DS2 has an extra evaluation set of solely pathological cases to test the 
accuracy of trained network. 

\begin{table}[t]
    \label{tab:dslengths}
    \centering
    \begin{tabular}{lcccc}
        Dataset & Sequence Length & Training & Validation & Evaluation \\
        \hline
        MLDS-DS1 & 1024 & 2048 & 512 & 64 \\
        MLDS-DS2 & 1024 & 2048 & 512 & 64*2 \\
        MLDS-DS3 & 256 & 4096 & 512 & 512 \\
        \hline
    \end{tabular}
    \caption{Training set details for all three MLDS datasets}
    \label{tab:my_label}
\end{table}

\subsection{Random Automata Generation}
MLDS-DS3 consists of networks which model machines based on randomly generated 
automata.  The algorithm for generating these graphs is listed in 
Algorithm~\ref{alg:rgraph}. 

\begin{algorithm}[H]
    \label{alg:rgraph}
    \SetAlgoLined
    \KwResult{A directed graph that represents a randomly-generated automaton}
    graph = create empty directed graph\;
    node\_array = create N nodes\;
    shuffle(node\_array)\;
    mark emitter nodes\;
    \For{node in node\_list}{
        graph = add node\;
    }
    
    \For{node in node\_list}{
        edge\_label = random\_choice(alphabet)\;
        graph = add\_edge(node, node+1, edge\_label)\;
    }
    
    \For{edge in graph}{
        n1 = original node\;
        n2 = destination node\;
        \For{label in other entries in alphabet}{
            new\_dest = random choice(node in graph and not n2)\;
            graph = add\_edge(n1, new\_dest, label)\;
        }
    }
    \caption{Generates automaton from an alphabet, a number of states, and emitter states.}
\end{algorithm}

\subsection{PyTorch Network Architecture}
\begin{figure}[t]
    \centering
    \includegraphics[width=0.5\textwidth]{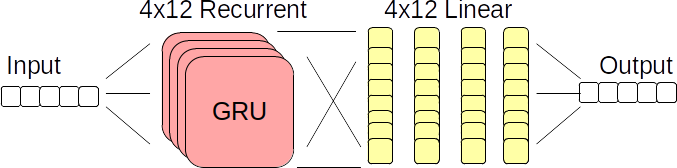}
    \caption{Network architecture for MLDS-DS1 and MLDS-DS2}
    \label{fig:ds12-network}
\end{figure}
\begin{figure}[t]
    \centering
    \includegraphics[width=0.5\textwidth]{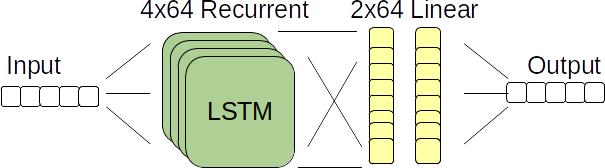}
    \caption{Network architecture for MLDS-DS3}
    \label{fig:ds3-network}
\end{figure}

MLDS-DS1/DS2 models each machine with a 4-layer deep GRU, followed by 4 
fully-connected linear levels, with no activation functions between each layer.
Each layer has a width of 12.  Pseudo-code for this is shown in 
Figure~\ref{fig:ds12-network}.  MLDS-DS3 uses exactly the same network, except
with 4-layer stacked LSTM and 2 fully connected layers with no activation 
functions.  However, each layer has a width of 64, leading to a significantly
more complex network. The code for DS3 is the same as in 
Figure~\ref{fig:ds3-network}, but with LSTM instead of GRU.

\end{document}